%% file: samplepaper.tex
\documentclass[runningheads]{llncs}
\usepackage{graphicx}
\usepackage{csquotes}
\usepackage{multirow}
\usepackage{makecell}
\usepackage{caption}
\usepackage{amsmath}
\usepackage{xcolor}
\usepackage{color,soul}
\usepackage{amssymb}
\usepackage{algorithm}
\usepackage[noend]{algpseudocode}
\usepackage{float}
\usepackage{multirow}

\graphicspath{ {./images/} }
\begin{document}
\title{Regularized Evolutionary Algorithm for\\ Dynamic Neural Topology Search}
%
%
\author{Cristiano Saltori\inst{1}, Subhankar Roy\inst{1,2}, Nicu Sebe\inst{1} \and Giovanni Iacca\inst{1}}
\authorrunning{C. Saltori et al.}
%
\institute{Dept. of Information Engineering and Computer Science, University of Trento, Italy \and
Fondazione Bruno Kessler (FBK), Trento, Italy \\
\email{cristiano.saltori@studenti.unitn.it,\\
\{subhankar.roy, niculae.sebe, giovanni.iacca\}@unitn.it}}
\maketitle              
\begin{abstract}
Designing neural networks for object recognition requires considerable architecture engineering. As a remedy, \textit{neuro-evolutionary} network architecture search, which automatically searches for optimal network architectures using evolutionary algorithms, has recently become very popular. Although very effective, evolutionary algorithms rely heavily on having a large population of individuals (i.e., network architectures) and are therefore memory expensive. In this work, we propose a Regularized Evolutionary Algorithm with low memory footprint to evolve a dynamic image classifier. In details, we introduce novel custom operators that regularize the evolutionary process of a micro-population of 10 individuals. We conduct experiments on three different digits datasets (MNIST, USPS, SVHN) and show that our evolutionary method obtains competitive results with the current state-of-the-art.

\end{abstract}

\input{intro.tex}

\input{related.tex}
\input{method.tex}
\input{experiments.tex}
\vspace{-0.1cm}
\section{Conclusions}
\vspace{-0.1cm}
In this work we have addressed Neural Architecture Search through a Regularized Evolutionary Algorithm. The proposed algorithm is especially useful when limited memory and limited computational resources are at disposal. Our main contributions can be summarized as follows: i) an evolving cell topology with variable number of hidden nodes; ii) ad hoc crossover and mutation operators for improved exploration; and iii) a novel stagnation avoidance for better convergence. We have conducted experiments on three different digits datasets and obtained competitive results. As a future work, we plan to scale our method to larger populations of individuals (parallelized on multiple GPUs) and test it on more complex datasets, also including videos.
\vspace{-0.2cm}
\section*{Acknowledgments}
\vspace{-0.3cm}
We gratefully acknowledge the support of NVIDIA Corporation with the donation of the TITAN Xp GPU used for this research.

%
%
%

\input{appendix.tex}

\end{document}

%% file: intro.tex
\section{Introduction}
\label{sec:intro}

Deep Learning has made a remarkable progress in a multitude of computer vision tasks, such as object recognition~\cite{krizhevsky2012imagenet,roy2019unsupervised}, object detection~\cite{ren2015faster}, semantic segmentation~\cite{he2017mask}, etc. The success of deep neural networks (DNNs) has been attributed to the ability to learn hierarchical features from massive amounts of data in an end-to-end fashion. Despite the breakthroughs of manually designed networks, such as Residual Networks~\cite{he2016deep} and Inception Net~\cite{szegedy2016rethinking}, these networks suffer from two major drawbacks: i) the skeleton of the architectures is not tailored for a dataset at hand; and ii) it requires expert knowledge to design high-performance architectures. To circumvent the cumbersome network architecture design process, there has been lately an increased interest in automatic design through \textit{Neural Architecture Search}~\cite{elsken2019neural} (NAS). The objective of NAS is to find an optimal network architecture for a given task in a data-driven way. In spite of being in its nascent stages, NAS has consistently outperformed hand-engineered network architectures on some common tasks such as object recognition~\cite{zoph2018learning}, object detection~\cite{zoph2018learning} and semantic segmentation~\cite{chen2018searching}.

Among NAS methods, \textit{neuro-evolutionary} algorithms \cite{real2017large,Yaman2018,real2018regularized,stanley2019designing} have recently resurfaced that use bio-inspired evolutionary principles for finding optimized neural architectures. A promising work in this direction was conducted by Real \textit{et al}.~\cite{real2018regularized}, who proposed an \textit{aging}-based evolutionary algorithm and demonstrated that favouring younger individuals (i.e., network architectures) against the best individuals in a population is beneficial in terms of convergence to better network architectures.

The performance of neuro-evolutionary methods is, however, heavily dependent on the population size (i.e, the number of individuals in the current population) and as a result on the availability of computational resources. For instance, Real \textit{et al}.~\cite{real2018regularized} used a population of 100 individuals evaluated in parallel on 450 GPUs. In this work, we propose an improved evolutionary algorithm, named \textit{Regularized Evolutionary Algorithm}, to accomplish NAS in limited computational settings. Specifically, our method is based on the regularized algorithm by Real \textit{et al}.~\cite{real2018regularized} with various modifications that can be summarized as follows: i) an evolving cell with a variable number of hidden nodes where each node (see Fig.~\ref{fig:arch}b) can be thought of as a pairwise combination of common operations (such as convolution, pooling, etc); ii) a custom \textit{crossover} operation that allows recombination between two parent individuals in a population, to accelerate the evolutionary search; iii) a custom \textit{mutation} mechanism, to generate new architectures by randomly modifying their hidden states, operations and number of hidden nodes inside a cell; and iv) a \textit{stagnation avoidance} mechanism to avoid premature convergence. %
More details on the algorithm can be found in the Supplementary Material available at \url{https://arxiv.org/abs/1905.06252}.
Noticeable is the implicit regularization performed by the stagnation avoidance and the mutation of $B$, as shown in details in Sec. \ref{sec:evolution}.

As we show in our experiments conducted on three different \textit{digits} datasets, the proposed modifications improve on the exploratory capability of the evolutionary algorithm when very limited memory and computational resources are at disposal during training. In our experiments, we used a micro-population of 10 individuals evaluated on a single GPU, obtaining competitive results when compared with much larger computational setups such as those used in \cite{real2018regularized}.

%% file: related.tex
\vspace{-0.2cm}

\section{Background}
\label{sec:rel}

The recent successes of NAS algorithms~\cite{bergstra2013making,domhan2015speeding,baker2016designing,zoph2017neural,zhong2018practical,zoph2018learning,real2018regularized,stanley2019designing} have opened new doors in the field of Deep Learning by outperforming traditional DNNs~\cite{krizhevsky2012imagenet,he2016deep} that are manually designed. Mostly, the NAS algorithms differ in the way they explore the search space of the neural architectures. The search strategy can be broadly divided into three different categories. The first category of methods~\cite{bergstra2013making,domhan2015speeding} use Bayesian optimization techniques to search for optimal network architectures and have led to some early successes before reinforcement learning (RL) based methods \cite{baker2016designing,zoph2017neural,zoph2018learning} became mainstream, which is the second predominant category. In RL based methods the agent's action can be considered as the generation of a neural architecture within a given search space, whereas the agent's reward is the performance of the trained architecture on the validation set. Zoph \textit{et al}.~\cite{zoph2017neural} used a recurrent neural network to sample a set of operations for the network architecture. In another work, the best performing network, dubbed as NASNet~\cite{zoph2018learning}, is searched by using Proximal Policy Optimization.

The final category of NAS algorithms leverage evolutionary algorithms for finding the optimal network architectures. Surprisingly, the first work on using genetic algorithm for searching network architectures \cite{miller1989designing} dates back several decades ago. Since then, there have been several works \cite{angeline1994evolutionary,stanley2002evolving} which use evolutionary algorithm both to search network architectures and to optimize the weights of the proposed networks. However, in practice Stochastic Gradient Descent (SGD) based optimization of network weights works better than evolution based optimization. Therefore, a series of methods \cite{real2017large,real2018regularized,stanley2019designing} have been proposed which restrict the usage of evolutionary algorithm just to guide architecture search, while using SGD for network weight optimization. The evolution based NAS methods differ in the way the sampling of parents are done, the update policy of the population, and how the offspring are generated. For instance, in \cite{real2018regularized} parents are selected according to a tournament selection \cite{goldberg1991comparative}, while in \cite{elsken2019efficient} this is accomplished through a multi-objective Pareto selection. The variations in the population update can be noticed in \cite{real2017large} and \cite{real2018regularized} where the former removes the worst individual and the latter removes the oldest individual from a population, respectively. Similar to \cite{real2017large}, but differently from \cite{real2018regularized}, in this work we propose a $\mu+\lambda$ methodology to remove the worst individuals from the population, where $\mu$ and $\lambda$ is the population size and number of offspring, respectively. With our approach, the best individual is preserved in the population (elitism). 
Furthermore, as opposed to \cite{real2017large,real2018regularized} who only considered mutation as the sole offspring generation process, here we additionally consider a crossover operator whose purpose is to accelerate the evolutionary process by recombining ``building blocks'' obtained by the parents undergoing crossover.

Finally, the choice of population size plays a very crucial role in the exploration of the search space of network architectures. For instance, the availability of mammoth computing power (450 GPUs) allowed \cite{real2018regularized} to use a population size $P = 100$, as opposed to our experiments which use a micro-population of size $P = 10$. Thus, we compensate the lack of computational resources with various modifications to the evolution process, so to allow an efficient exploration of the search space with a very limited computational budget.

%% file: method.tex
\vspace{-0.3cm}

\section{Method}
\label{sec:label}

In this section we present our method by first introducing the search space, and then describing in detail the proposed Regularized Evolutionary Algorithm.

\vspace{-4mm}
\subsection{Search space}
\label{sec:search}
We consider the NASNet search space, introduced in \cite{zoph2018learning}, comprised of image classifiers that have a fixed structure (see Fig.~\ref{fig:arch}a), each one composed of repeated motifs called \textit{cells}, similar to Inception-like modules in \cite{szegedy2016rethinking}. Each architecture is composed of two building-block cells called \textit{normal cell} and \textit{reduction cell}. By design, all the normal and reduction cells should have the same architecture, but the internal structure of a normal cell could vary from the reduction cell (see Sec.~\ref{sec:results}). Another fundamental difference between the two is that the reduction cell is followed by an average pooling operation with a $2\times2$ kernel and stride of 2, whereas the normal cells preserve the size.
\vspace{-0.6cm}
\begin{table}[!ht]
    \centering
    \begin{tabular}{cc}
         \includegraphics[width=0.18\textwidth]{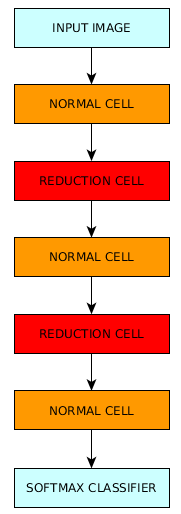} & \includegraphics[width=0.80\textwidth]{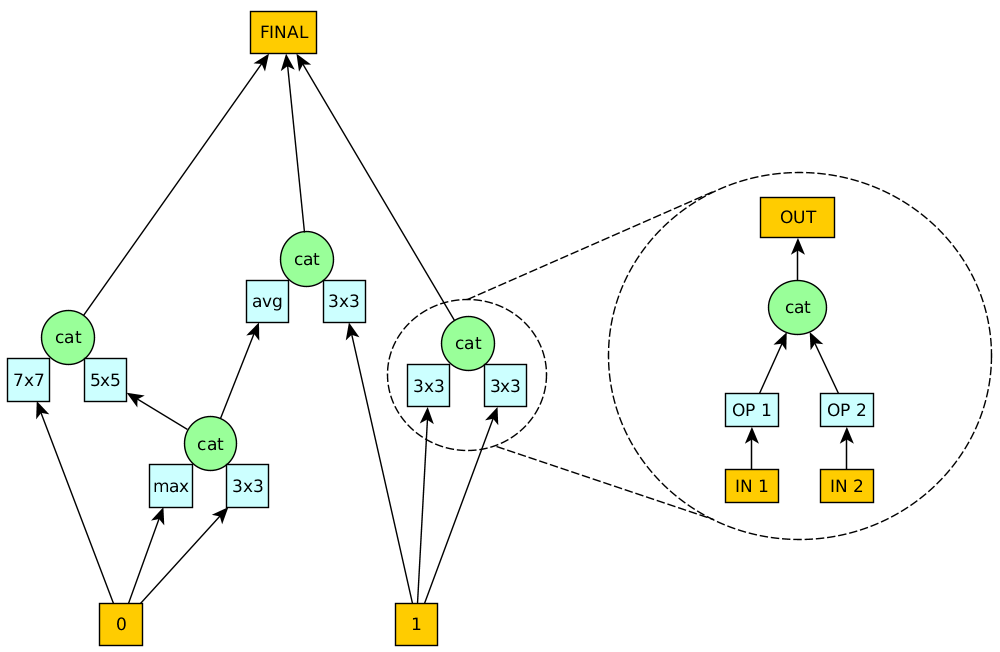} \\
         (a) & (b)         
    \end{tabular}
    \captionof{figure}{(a) The full outer architecture used during the search phase; (b) Example of an evolved cell's internal structure and the depiction of a hidden node's structure.}
    \label{fig:arch}
\end{table}
\vspace{-1cm}

As seen from Fig.~\ref{fig:arch}b and borrowing the taxonomy from \cite{zoph2018learning}, each cell is composed of two input states (designated as \enquote{0} and \enquote{1}), an output state (designated as \enquote{final}) and a variable number of $B$ hidden nodes. Excluding the input and output states, each \textit{intermediate} hidden node is constructed by a pairwise combination of two independent operations (or \textit{op}) on two inputs, followed by the concatenation of their corresponding outputs. We refer to each hidden node's output as hidden states.

In total, each cell has a maximum of $B+3$ states. Ops consist of commonly used convnet operations, essentially only a subset of \cite{zoph2018learning}, defined as: identity layer; $3\times3$ convolution; $5\times5$ convolution; $7\times7$ convolution; $3\times3$ max pooling; $3\times3$ average pooling.
For example, in Fig.~\ref{fig:arch}b, the \textit{zoomed-in} hidden node is constructed by two unique 3 $\times$ 3 convolution operations on the input state \enquote{1}, followed by concatenation of resulting feature maps to yield another hidden state. It is worth noting that in \cite{real2018regularized,zoph2018learning} $B$ is fixed for both the normal and reduction cell, while in our method we treat $B$ as a parameter which is also optimized during the evolution, thus allowing different $B$ (where $B \in [B_{min}, B_{max}]$) for normal and reduction cells. In details, the input state is composed of a $1\times1$ convolutional layer with stride 1, to preserve the number of input channels, while the output state is composed of a concatenation of the activations coming from the previous unused hidden states. 
The inputs to each intermediate hidden node can come from any of the previous states, including duplicates, while the output from a hidden node can be connected to any of the following states. Finally, the hidden states that are not connected with any (unused) op are concatenated in the output hidden state.

Given a normal and a reduction cell, the architecture is built by alternating these motifs as in \cite{real2018regularized}, with the major difference that we avoid the use of skip-connections and thus reduce the number of repetitions, in order to obtain a simpler and faster training procedure.
As in \cite{real2018regularized}, we keep the same number of feature maps inside each cell fixed, while this number is multiplied by a factor $K$ after each reduction cell. A sample overall architecture structure is reported in Fig.~\ref{fig:arch}a. The final goal of the NAS process is to find optimal architectures for normal and reduction cells. Once their architectures have been chosen by the algorithm, there are two hyper-parameters that need to be determined: the number of normal cells ($N$) and the number of output features or filters ($D$).
\vspace{-0.2cm}

\subsection{Regularized Evolutionary Algorithm}
\label{sec:evolution}
\noindent The proposed Regularized Evolutionary Algorithm, see Algorithm \ref{alg:evolution}, is based on a micro-population of $P$ = 10 individuals (i.e., network architectures). Each individual represents a solution in the NASNet space, which as said is further composed of normal and reduction cells. During evolution the population is evolved in order to maximize the classification accuracy on a hold out validation set. In the following, we describe in details each step of the proposed algorithm. 

\noindent\textbf{Population initialization}. In this initial step $P$ individuals with random cells are initialized and evaluated. Since we use a small value of $P$, the population is strongly affected by badly initialized solutions. To solve this problem we initialize all individuals with $B$ hidden nodes (with $B$ randomly selected in the range [$B_{min}$, $B_{max}/2$]), in order to allow the algorithm to optimize and gradually increase $B$ during the evolution.

\noindent\textbf{Offspring generation}. At each generation $F$ offspring (i.e., new network architectures) are generated and inserted into the population. More specifically, for each i-th offspring to generate, 1 $\leq$ $i \leq F$, a sample of $S$ individuals is selected randomly from the population $P$. From these $S$ individuals, two are selected as parents for the i-th offspring: the best individual in the sample, $P_1$ (this is equivalent to a tournament selection), and another individual $P_2$, which is instead randomly selected from the sample. $P_1$ and $P_2$ are then used to generate the offspring through crossover and mutation.

\noindent\textbf{Crossover}. The crossover operation aims at merging two parent individuals, thus enabling inheritance through generations. Let us assume that $P_1$ and $P_2$ have $B_1$ and $B_2$ hidden nodes, respectively. The offspring cell, derived from $P_1$ and $P_2$, will have $B_{temp} = max(B_1, B_2)$ intermediate hidden nodes. The hidden nodes of the parent cells are then merged through sequential random selection with probability $\tau_{cross}$, as follows. For the sake of generality we assume $B_1 \neq B_2$. The first $min(B_{1}, B_2)$ intermediate hidden nodes $H^{off}_j$ in the offspring network, 1 $\leq$ $j \leq min(B_{1}, B_2)$, are set as:
\[
H^{off}_j =
\begin{cases}
H^{P_1}_{j} & \text{if $p \leq \tau_{cross}$} \\
H^{P_2}_{j} & \text{otherwise} 
\end{cases}
\]
where $p$ is a uniform random number drawn in $[0,1)$ and $\tau_{cross}$ is the crossover probability threshold. This threshold plays an important role in the exploration/exploitation balance in the algorithm, allowing to inherit the hidden nodes either from $P_1$ if $p$ $\leq \tau_{cross}$ (more exploitation), or from $P_2$ if $p > \tau_{cross}$ (more exploration). In our experiments, we set $\tau_{cross}=0.6$ to have a higher probability of inheriting the hidden nodes from the best individual in the sample ($P_1$), rather than from a random individual $P_2$. The remaining intermediate hidden nodes $H^{off}_j$, $min(B_{1}, B_2) < j \leq B_{temp}$, are then derived directly from the corresponding j-th node in the parent which has higher $B$. The edges of each hidden node in the offspring are inherited from the parent without modifications.

\noindent\textbf{Mutation}. The mutation operation is a key ingredient in the evolution process, as it allows the search to explore new areas of the search space but also refine the search around the current solutions. Apart from the \textit{op mutation} and \textit{hidden state mutation} introduced in \cite{real2018regularized}, here we additionally introduce a specific mutation for tuning the number of hidden nodes $B$. The first step is the op mutation which requires a random choice to be made in order to select either the normal cell or the reduction cell $C$. Once selected, the mutation operation will select one of the $B$ pairwise combinations (or hidden nodes) at random. From the chosen pairwise combination, one of the two ops is replaced, with a probability $\tau_{m-op}$, by another op allowed in the search space. Following op mutation, hidden state mutation is performed which, akin to op mutation, also oversees random choice of cells, one of the $B$ pairwise combinations and one of the two elements in a combination. However, hidden state mutation differs from op mutation in that it replaces (with probability $\tau_{m-edge}$) the incoming hidden state or cell input, which corresponds to the chosen element, with another one that is inside $C$, instead of the op itself. It is important to notice that the new input is chosen from the previous hidden nodes or cell inputs to ensure the feed-forward nature of a cell.
Finally, to encourage further exploration a third mutation allows to increment the parameter $B$ of $C$ by 1, with probability $\tau_b$, such that a new node with index $B+1$ is introduced with random weights, if $B < B_{max}$. It is to be noted that all unused states are concatenated at the end of the cell. Finally, the offspring individual is evaluated and added to the population.

\noindent\textbf{Survivor selection}. The survivor selection is conducted at the end of each generation. It is crucial since it permits the removal of the worst solutions from the population and facilitates the convergence of the evolutionary search. In this phase, the selection and removal of the worst individuals is conducted according to a $\mu+\lambda$ selection scheme, where based on the symbols we have used so far $\mu=P$ is the population size and $\lambda=F$ is the number of generated offspring. Therefore, at each step $F$ worst individuals are removed from the set of $P+F$ individuals, so that the parent population for the next generations has a fixed size of $P$ individuals. It should be noted that, compared to other survivor selection schemes, this scheme is implicitly elitist (i.e., it always preserves the best individuals found so far) and also allows inter-generational competition between parents and offspring.

\noindent\textbf{Stagnation avoidance}. To avoid premature convergence of the population to a local minimum we propose two stagnation avoidance mechanisms: \textit{soft} and \textit{hard} stagnation avoidance. When stagnation occurs (i.e., the best accuracy does not improve for more than $A_{stag}$ generations, as detected in the method StagnationDetected() in Algorithm \ref{alg:evolution}), the mutation probabilities $\tau_{m-op}$ and $\tau_{m-edge}$ are increased to $\tau_{m-op-avoid}$ and $\tau_{m-edge-avoid}$, respectively, where $\tau_{m-op-avoid}$ and $\tau_{m-edge-avoid}$ denote increased mutation probabilities (such that $\tau_{m-op-avoid} > \tau_{m-op}$ and $\tau_{m-edge-avoid} > \tau_{m-edge}$). The choice to increase $\tau_{m-op}$ and $\tau_{m-edge}$ is due to the fact that higher mutation probabilities, in general, allow a higher exploration of the search space. On the other hand, $\tau_b$ is not increased because the addition of more hidden nodes (which would likely follow from a higher mutation probability) would increase the complexity of the cell and does not alleviate stagnation. Finally, the hard stagnation avoidance is a brute force remedy to stagnation. Specifically, if soft stagnation avoidance fails, $P-1$ worst solutions are replaced by $P-1$ individuals with random cells, while retaining the best solution found so far.
\vspace{-0.6cm}
\begin{algorithm}[!ht]
\caption{Regularized Evolutionary Algorithm}
\label{alg:evolution}
\begin{algorithmic}
\State $population$ $\leftarrow$ $\emptyset$, $softTried$ $\leftarrow$ False, $g$ $\leftarrow$ 1\;
\While{$|population|$ $<$ $P$} \Comment{Initialize population}
    \State $individual.arch$ $\leftarrow$ RandomArchitecture()
    \State $individual.accuracy$ $\leftarrow$ TrainAndEval($individual.arch$)
    \State add $individual$ to $population$
\EndWhile  \label{init}
\While{$g \leq G$} \Comment{Evolve for $G$ generations}
    \While{$|population|$ $<$ $P+F$} \Comment{Generate $F$ offspring}
        \State $sample$ $\leftarrow$ $\emptyset$ \Comment{Parent candidates}
        \While{$|sample|$ $<$ $S$}
            \State $candidate$ $\leftarrow$ randomly sampled $individual$ from $population$
            \State add $candidate$ to $sample$
        \EndWhile
        \State $P_1$ $\leftarrow$ highest accuracy individual in $sample$  \Comment{Tournament selection}
        \State $P_2$ $\leftarrow$ randomly selected individual in $sample$
        \State $offspring.arch$ $\leftarrow$ Crossover($P_1$, $P_2$)
        \State $offspring.arch$ $\leftarrow$ Mutation($offspring.arch$)
        \State $offspring.accuracy$ $\leftarrow$ TrainAndEval($offspring.arch$)
        \State add $offspring$ to $population$
    \EndWhile
    \State remove $F$ worst individuals from $population$ \Comment{$\mu+\lambda$ selection}
    \If{StagnationDetected()} \Comment{Prevent Stagnation}
        \If{$softTried$} \Comment{Hard Stagnation Avoidance}
            \State remove $P-1$ worst individuals from $population$
            \State add $P-1$ random individuals to $population$
        \Else \Comment{Soft Stagnation Avoidance}
            \State $\tau_{m-op}$ $\leftarrow$ $\tau_{m-op-avoid}$, $\tau_{m-edge}$ $\leftarrow$ $\tau_{m-edge-avoid}$
            \State $softTried$ $\leftarrow$ True\;
        \EndIf
    \EndIf
    \State $g$ $\leftarrow$ $g+1$\;
\EndWhile
\end{algorithmic}
\end{algorithm}

%% file: experiments.tex
\vspace{-1.4cm}
\section{Experiments results}
\label{sec:exp}
\vspace{-0.5cm}
In this section we describe the datasets and report the experimental setup used in the evolutionary algorithm and network training. Finally, we report our experimental evaluation on the considered datasets.

\vspace{-0.5cm}
\subsection{Datasets}
\label{sec:datasets}
We conducted all our experiments on the well-known MNIST, USPS and SVHN datasets, all consisting of digits ranging from 0 to 9. MNIST and USPS are taken from U.S. Envelopes and consist of grayscale handwritten digits. The SVHN is a real-world dataset of coloured digits taken from Google Street View.

\vspace{-3mm}
\subsection{Experimental setup}
\label{sec:exp_setup}
As mentioned in Sec.~\ref{sec:rel}, our proposed method belongs to the category of neuro-evolutionary methods where the optimal network is searched by an evolutionary algorithm while the weights of the networks are trained with gradient descent algorithm. Hence, we separately provide details about the chosen experimental setup for the evolutionary algorithm and the network architecture and training. For the experiments we used a Linux workstation equipped with a CPU Intel(R) Core(TM) i9-7940X and a TITAN Xp GPU. The lack of massive computation resource has led us to the following choice of parameters.

\noindent\textbf{Evolutionary algorithm}. The experiments were conducted with $P=10$, $F=10$, $B_{min}=2$, $B_{max}=6$ and ran for $G=200$ generations. A sample size $S=2$ has been used to have a low selection pressure. For the crossover operation, $\tau_{cross}=0.6$ was used to allow a higher probability of exploitation of the best solutions in the populations. Instead, in the mutation phase, the probabilities $\tau_{m-op}$, $\tau_{m-edge}$ and $\tau_b$ were set to 0.4, 0.4 and 0.2, respectively, to allow both a moderate exploration of the search space without significantly increasing the complexity of the cells. Finally, the stagnation check has been enabled after 50 generations and the evolution is considered to be stagnating if the same best result (in terms of accuracy) is provided for at least $A_{stag}= 40$ generations. In that case, $\tau_{m-op-avoid}$ and $\tau_{m-edge-avoid}$ were both set to 0.6.

\noindent\textbf{Network architecture}. The network architectures which conform to the overall structure in Fig.~\ref{fig:arch}a were trained during the experiments. Due to limited memory, the initial channels $D$ in the normal cell was set to 24 and then subsequently multiplied by a factor $K=2$ after each reduction cell, with a total of 3 normal cells and 2 reduction cells. During the search phase, each individual has been trained for 15 epochs with mini-batches of size 64 using an Adam optimizer having an initial learning rate of 1e-4 with exponential decay. Dropout layer with drop probability of 0.2 has been used in the final softmax classifier. To further reduce the computations, our cells take duplicated output states $H_{prev}$ from the previous cell instead of $H_{prev}$ and $H_{prev - 1}$, thereby eliminating skip connections between cells unlike \cite{zoph2018learning,real2018regularized}. At the end of the search phase, when the final best individual has been retrieved by the evolutionary algorithm, a training has been performed with that best individual for 100 epochs with the same previous parameters but with an initial learning rate of 1e-3. Batch normalization layers were also inserted after each convolutional layer. 
Standard cross-entropy loss was used for training the networks.

\vspace{-3mm}
\subsection{Results}
\label{sec:results}

Since the exploration in the NASNet search space is time consuming and also memory intensive, the best cells were searched only using the MNIST dataset due to its small size and reduced complexity. This has led to the evolution of a best normal cell and a best reduction cell, which collectively form the best network, dubbed as \textit{EvoA}. The second best network is called \textit{EvoB}.

To show the benefit of having variable $B$ hidden nodes in a cell, we considered the following baselines: i) EvoA, our best network with variable $B$; and ii) EvoB, our second best network also with variable $B$; iii) Model-A with $B=3$; and iv) Model-B with $B=4$. As seen from Tab.~\ref{tab:ablation}, EvoA, the network found by our proposed method, outperforms all baselines which consider fixed $B$. This highlights the advantage of keeping $B$ variable.

\vspace{-0.7cm}
\begin{table}[!ht]
    \centering
    \begin{tabular}{cc}
         \includegraphics[width=0.4\textwidth, height=0.4\textwidth]{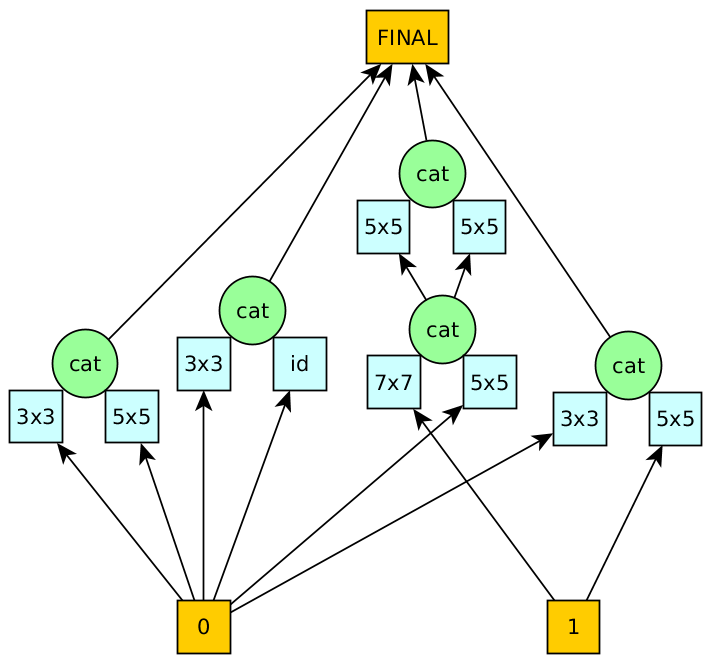} & \includegraphics[width=0.45\textwidth, height=0.4\textwidth]{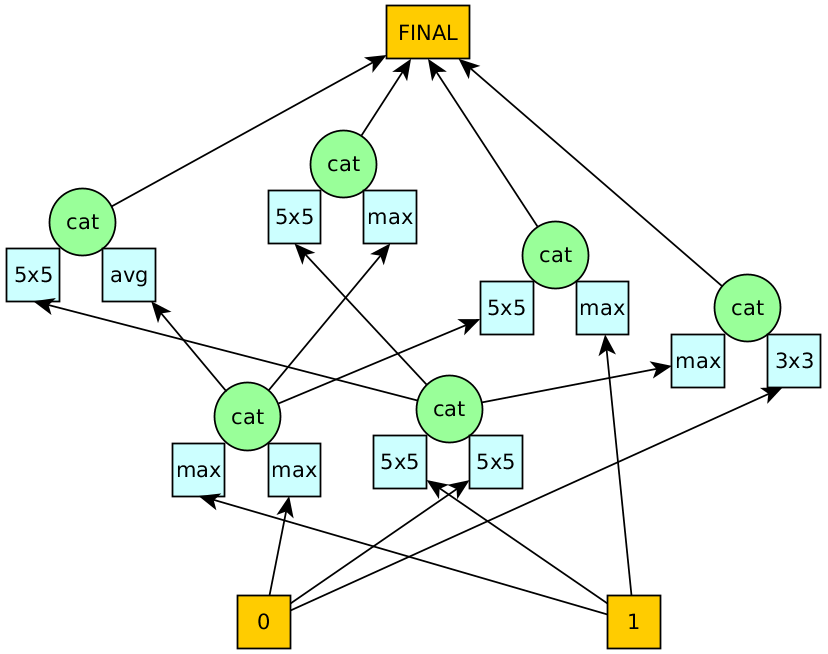} \\
         (a) & (b)         
    \end{tabular}
    \captionof{figure}{Architecture of the best cells of EvoA found using MNIST: (a) Normal cell; (b) Reduction cell. The inputs \enquote{0} and \enquote{1} are the output states from the previous cells. The output \enquote{FINAL} is the concatenation operation resulting from the remaining hidden nodes inside the cell. The blocks in cyan represent a pairwise combination of operations and their concatenation. Note that the colour corresponds to Fig.~\ref{fig:arch}b.}
    \label{fig:evoa}
\end{table}
\vspace{-1.6cm}
\begin{table}[!h]
    \centering
    \caption{Ablation study comparing the classification accuracy of our best evolved networks EvoA and EvoB with networks having fixed $B$ hidden nodes inside cells.}
    \setlength{\tabcolsep}{5pt}
    \begin{tabular}{lcccc}
        \hline
        Model & EvoA & EvoB & Model-A & Model-B\\
        \hline
        Hidden States & Variable & Variable & 3 & 4\\
        Final Accuracy & \textbf{99.59} & 99.55 & 99.47 & 99.49\\ 
        \hline
    \end{tabular}
    \label{tab:ablation}
\end{table}

\vspace{-1.2cm}
\begin{table}[!ht]
    \caption{Digits test set results for EvoA and EvoB when compared with baselines.}
    \centering
    \setlength{\tabcolsep}{5pt}
    \begin{tabular}{l|cc|cc}
        \hline
        \multirow{2}{*}{Model} & \multicolumn{2}{c|}{MNIST} & \multicolumn{2}{c}{SVHN}\\
         & \# Parameters & Accuracy (\%) & \# Parameters & Accuracy (\%)\\ \hline
        ResNet18~\cite{he2016deep} & 11.18M & 99.56 & 11.18M & 92.00 \\
        DeepSwarm~\cite{pattio2019deep} & 0.34M & 99.53 & 0.34M & \textbf{93.15}\\
        EvoB (ours) & 0.21M & 99.55 & 0.37M & 91.56 \\
        EvoA (Ours) & 0.22M & \textbf{99.59} &  0.39M & 91.80\\
        \hline
    \end{tabular}
    \label{tab:digits}
\end{table}
\vspace{-0.6cm}
In Fig.~\ref{fig:evoa} the architectures of EvoA's internal cells, both normal and reduction, are presented. As it can be observed, normal cells in EvoA have $B$=5 hidden nodes whereas the reduction cell has $B$=6. While the input states \enquote{0} and \enquote{1} are duplicates of each other, unlike \cite{zoph2018learning,real2018regularized}, we still observed good performance without using skip connections.

We have compared our method with the following baselines: i) ResNet18~\cite{he2016deep} and ii) DeepSwarm~\cite{pattio2019deep}, a NAS algorithm based on Ant Colony Optimization. Due to the lack of availability of code from our closest competitor \cite{real2018regularized}, we re-implemented their aging-based evolution but the solutions never converged due to the small size of our population, caused by limited memory and GPU availability. This resonates the fact that \cite{real2018regularized} is feasible only when large compute resources are available. As seen from Tab.~\ref{tab:digits}, EvoA outperforms all baselines on MNIST, with the lowest number of tunable parameters. Importantly, for the SVHN experiments we used an augmented architecture (with $N=4$, $D=32$ and each normal cell repeated thrice), composed of already evolved cells obtained from the search phase with MNIST. The network, despite not being evolved with SVHN, produced competitive results (slighly worse than ResNet18 and DeepSwarm).

\vspace{0.1cm}

\noindent\textbf{Transfer learning}. We also investigated the transferability of the trained networks on unseen target data. In details, we trained EvoA and EvoB on MNIST and used the trained weights to classify USPS test images. As seen from Tab.~\ref{tab:usps}, our EvoA and EvoB networks outperform all the baselines. Notably, while DeepSwarm outperforms both EvoA and EvoB on SVHN (as seen in Tab. \ref{tab:digits}), it performs significantly worse when used for transfer learning. On the other hand, our networks achieve near target test accuracy on unseen data.
\vspace{-10mm}
\begin{table}[!ht]
    \centering
    \setlength{\tabcolsep}{5pt}
    \caption{Classification accuracy on USPS with networks trained only on MNIST.}
    \begin{tabular}{lcccc}
        \hline
        Model & ResNet18~\cite{he2016deep} & DeepSwarm~\cite{pattio2019deep} & EvoB (Ours) & EvoA (Ours)\\ \hline
        Accuracy (\%) & 89.08 & 61.56 & \textbf{96.72} & 96.40\\
        \hline
    \end{tabular}
    \label{tab:usps}
\end{table}
\vspace{-1cm}

%% file: appendix.tex
\section*{Supplementary Material}

\subsection*{Supplement A}
\label{app:A}
In this supplement, we further explain the novel operations introduced in Sec. \ref{sec:evolution} also including explanatory images.

\subsubsection*{Crossover}
\begin{figure*}[!ht]
    \centering
    \includegraphics[scale=0.38]{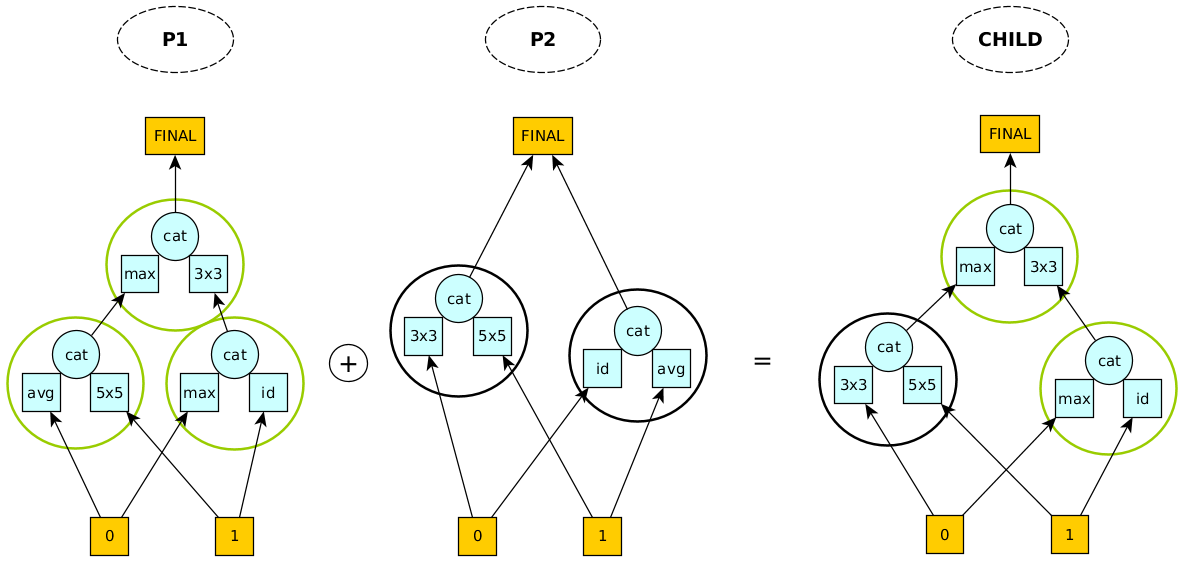}
    \caption{Example of crossover between P1 and P2. Green nodes are from P1 while black nodes come from P2.}
    \label{fig:crossover_example}
\end{figure*}

The crossover operator allows to exploit the existing solutions in the population. As previously said, we crossover two individuals by merging them at node level then inheriting the interconnections from the selected parents. Let us assume P1 and P2 are the best and randomly selected parents in Fig. \ref{fig:crossover_example}. As it can be seen, the parent P1 has $B=3$ while P2 has $B=2$ so, since $max(3, 2) = 3$, the resulting child will have $B=3$. At this point, the firsts two nodes, since $min(3, 2) = 2$, are randomly selected between P1 and P2 with $\tau_{cross}$ while the last node is selected from P1.
Finally, each node inherits the interconnections from the origin cell, see Fig. \ref{fig:crossover_example}.

\subsubsection*{Operation mutation}
\begin{figure*}[!ht]
    \centering
    \includegraphics[scale=0.4]{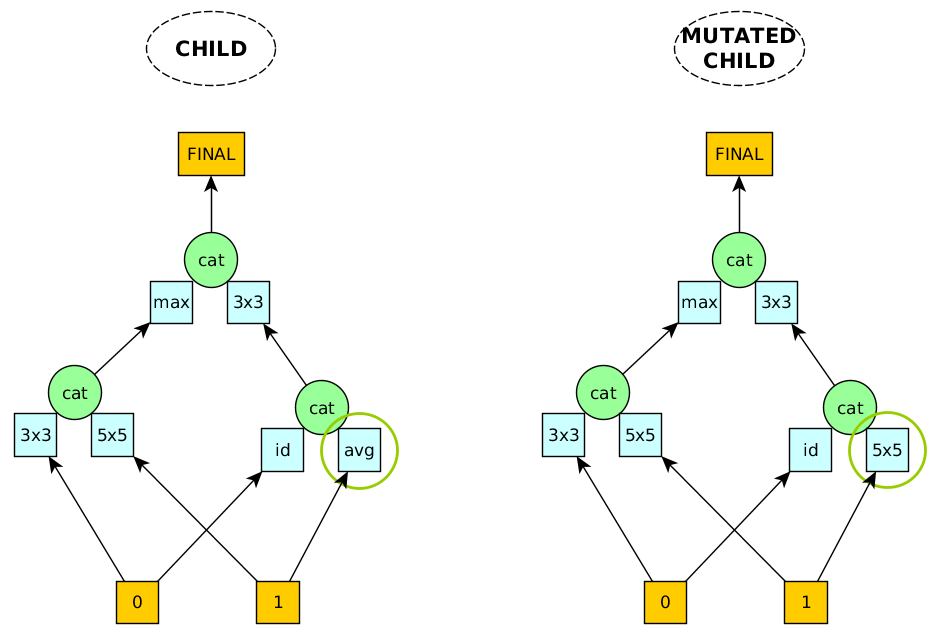}
    \caption{Example of operation mutation. The mutated operation is highlighted in green.}
    \label{fig:op_mutation_example}
\end{figure*}

The operation mutation consists in mutating a random operation in one of the child cells.
Starting from the child cell in Fig. \ref{fig:op_mutation_example}, that has been randomly selected between normal and reduction cells, firstly a random node is selected (in this case the bottom right node among all the child nodes). Secondly, a random operation is selected as a replacement. In the example, the average pooling op has been selected in the previously extracted node, see Fig. \ref{fig:op_mutation_example} highlighted in green. Finally, the operation is substituted with another randomly picked operation, in this case a $5\times5$ convolution, from the allowed ops in the search space.

\subsubsection*{Hidden State mutation}
\begin{figure*}[!ht]
    \centering
    \includegraphics[scale=0.42]{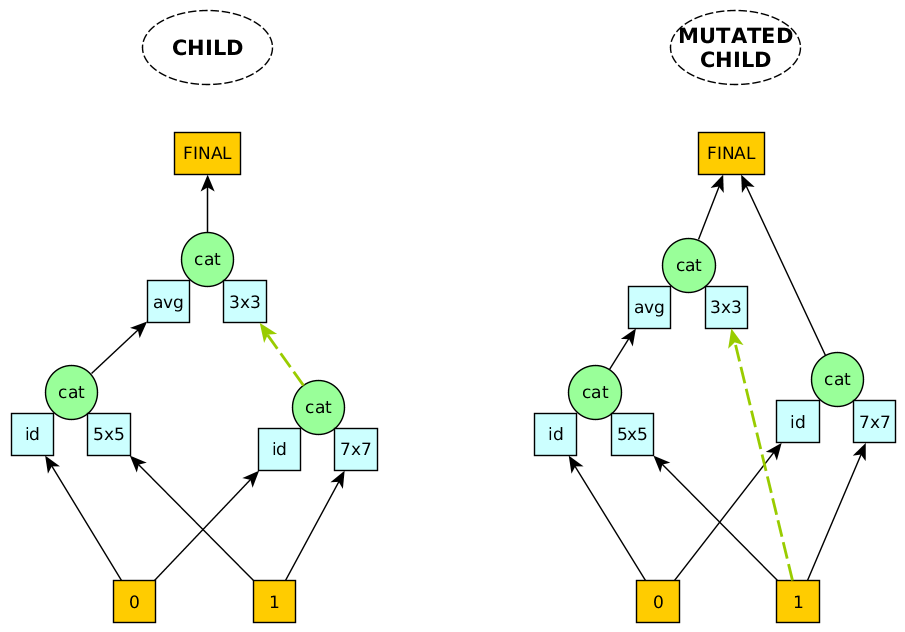}
    \caption{Example of hidden state mutation. The mutated connection is highlighted in dashed green.}
    \label{fig:edge_mut_example}
\end{figure*}

The hidden state mutation is similar to the operation mutation. Here, instead of mutating an operation, we mutate a connection between nodes. For the sake of generality, let us assume that the child in Fig. \ref{fig:edge_mut_example} is the obtained child from the previous crossover and mutation steps. At this point, one node is randomly selected among the $B=3$ nodes of the child. Finally, one of the input connection of the selected node is moved to another previous node. In the example, the top node is first selected and finally the right input connection is moved from the right node to the input state cell 1.

\subsubsection*{B mutation}
\begin{figure*}[!ht]
    \centering
    \includegraphics[scale=0.4]{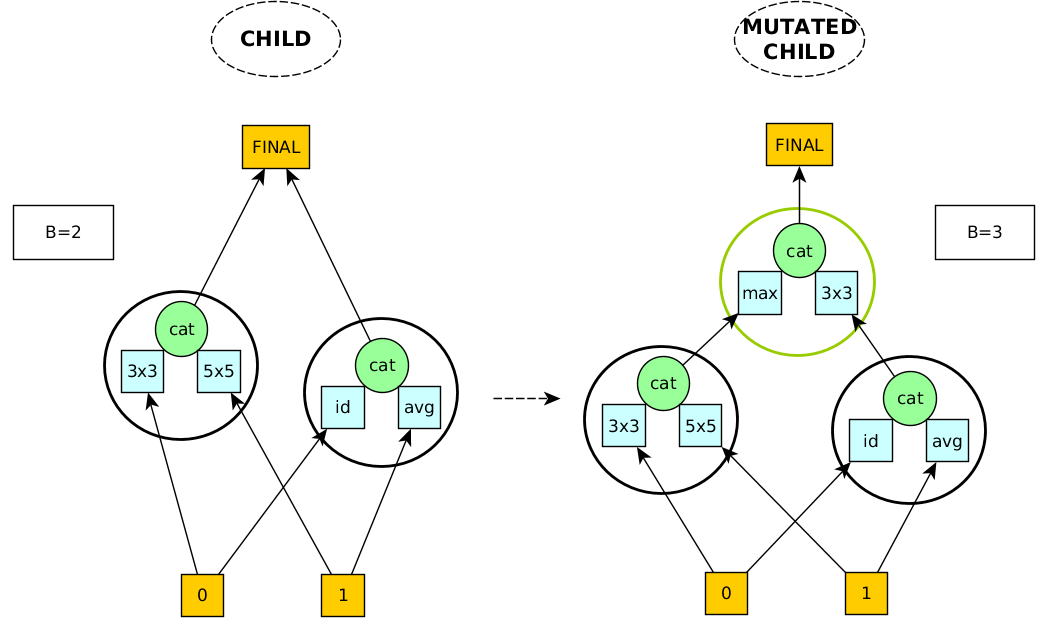}
    \caption{Example of B mutation. The new added node is highlighted by the green circle.}
    \label{fig:b_mutation_example}
\end{figure*}

The mutation of B consists in appending a hidden node to the child cell thus moving the child from having $B$ to $B+1$ hidden nodes.
Let us assume the child to be mutated is the one reported in Fig. \ref{fig:b_mutation_example} with an initial $B=2$.
After the mutation of B, an additional hidden node has been appended, highlighted in green in the example, to the previous nodes with random connections. At this point, the mutated child has a final $B = 3$.

\clearpage
\newpage
\pagebreak

\subsection*{Supplement B}
\label{app:B}

In this supplement we provide the internal structures of EvoB, the second best architecture evolved as reported in Sec. \ref{sec:results}.
The cells reported in Fig. \ref{fig:evob} have been obtained following the same procedure of EvoA.

\begin{table}
    \centering
    \begin{tabular}{cc}
         \includegraphics[width=0.49\textwidth, height=0.45\textwidth]{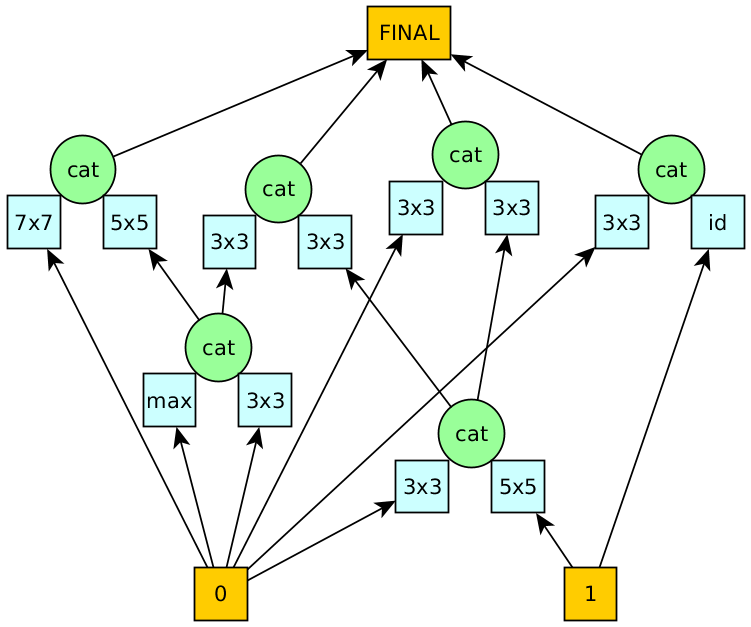} & \includegraphics[width=0.48\textwidth, height=0.45\textwidth]{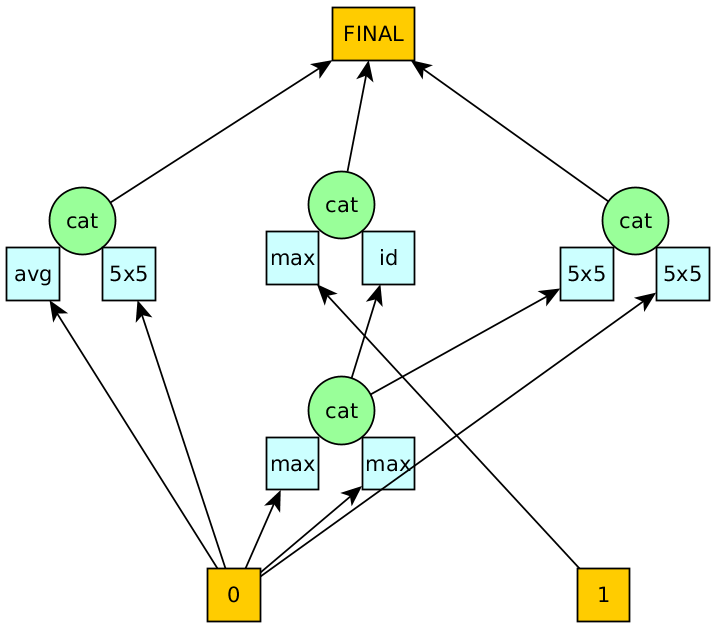} \\
         (a) & (b)         
    \end{tabular}
    \captionof{figure}{Architecture of the best cells of EvoB found using MNIST: (a) Normal cell; (b) Reduction cell. The inputs \enquote{0} and \enquote{1} are the output states from the previous cells. The output \enquote{FINAL} is the concatenation operation resulting from the remaining hidden nodes inside the cell. The blocks in cyan represent a pairwise combination of operations and their concatenation. Note that the colour corresponds to Fig.~\ref{fig:arch}b.}
    \label{fig:evob}
\end{table}